%% file: igarss_stereo.tex
\documentclass{article}
\usepackage{spconf,amsmath,epsfig}
\usepackage[utf8]{inputenc}

\usepackage{booktabs}
\usepackage{multirow}
\usepackage{xspace}
\usepackage{url}
\usepackage{comment}

\usepackage[skip=5pt]{caption}

\graphicspath{{./}}

\usepackage{color}
\definecolor{myGreen}{rgb}{0.2,0.6,0.}
\newcommand{\todo}[1]{{\color{red}TODO: #1 \color{black}}}

\newcommand{\RNum}[1]{{\bf (\lowercase\expandafter{\romannumeral #1\relax})}}
\newcommand{\wrt}{w.r.t.\@\xspace}
\newcommand{\cf}{cf.\@\xspace}

\newcommand{\ie}{i.e.\@\xspace}

\newcommand{\eg}{e.g.\@\xspace}
\newcommand{\Fig}{Fig.\@\xspace}

\newcommand{\myparagraph}[1]{\vspace{0.5em}\noindent\textbf{#1}}

\title{Self-supervised learning for stereo reconstruction on aerial images}
\name{Patrick Knöbelreiter \qquad Christoph Vogel \qquad Thomas Pock \vspace{-0.5cm}}
\address{\small \texttt{\{knoebelreiter, vogel, pock\}@icg.tugraz.at}\\ Institute for Computer Graphics and Vision\\Graz University of Technology}

\begin{document}
%

\setlength{\textfloatsep}{8pt}
\setlength{\floatsep}{9pt}

\maketitle
\begin{abstract}
\input{abstract}
\end{abstract}
%

\section{Introduction}
\label{sec:intro}
\input{intro}

\begin{figure}
  \centering
  \includegraphics[trim={0px 80px 50px 105px}, clip, width=\columnwidth]{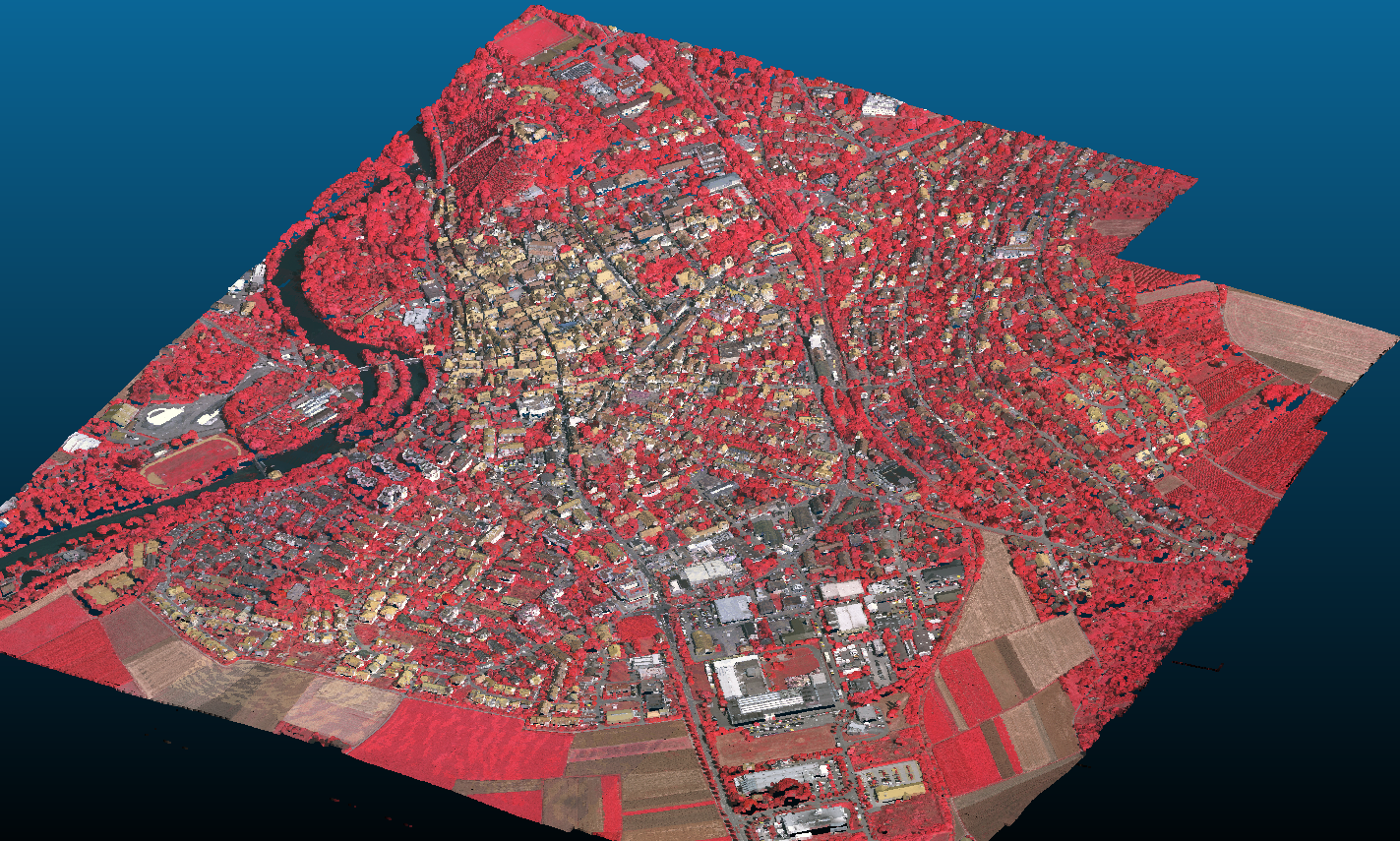}
  \caption{Visualization of the textured 3D point cloud of Vaihingen generated by our algorithm.}
\end{figure}
\section{Related Work}
\label{sec:relatedWork}
\input{related}
\section{Self-Supervised Dense Matching}
\label{sec:selfLearning}
\input{method}
\section{Experiments}
\label{sec:experiments}
\input{experiments}
\section{Conclusion \& Future Work}
\label{sec:conclusion}
\input{conclusion}


\small
\bibliographystyle{IEEEbib}
\bibliography{main,cnn-crf,ssvm,suppl,stereo}

\end{document}

%% file: abstract.tex
Recent developments established deep learning as an inevitable tool
to boost the performance of dense matching and stereo estimation.
%
%
On the downside, learning these networks requires a substantial amount of
training data to be successful.
Consequently, the application of these models outside of the 
laboratory is far from straight forward. 
In this work we propose a \emph{self-supervised training} procedure that allows
us to adapt our network to the specific (imaging) characteristics of the dataset
at hand, without the requirement of external ground truth data.
We instead generate interim training data by running our intermediate network
on the whole dataset, followed by conservative outlier filtering.
Bootstrapped from a pre-trained version of our hybrid CNN-CRF model, we
alternate the generation of training data and network training.
With this simple concept we are able to lift the completeness
and accuracy of the pre-trained version significantly.
We also show that our final model compares favorably to other popular
stereo estimation algorithms on an aerial dataset.
%

%% file: intro.tex
%
Acquired with modern high resolution cameras, aerial images can 
provide accurate 3D measurements of the observed scene via dense image matching.
%
Consequently, through the years, stereo estimation has emerged as an attractive
alternative to LiDAR 
(Light Detection and Ranging) in various tasks, like high resolution
Digital Surface Model (DSM) generation or orthoimage production~\cite{Gehrke_2012},
leading to a simplified processing pipeline and reduced (flight) costs.
Furthermore, the generation of stereo data is a common first step in many
different applications, 
\eg in 3D change detection \cite{Qin_2016} or
semantic 3D reconstruction of urban scenes \cite{BlahaVRWPS16}. 

Recently, machine learning, and in particular, deep learning has affected many
low-level vision tasks including stereo estimation, leading to considerable improved performance. 
Here, convolutional neural networks (CNNs) can be used to replace different parts
in conventional stereo pipelines, \eg the feature generation for computing the
data term~\cite{Zbontar2016}.
A different path is to directly formulate stereo estimation as a regression task~\cite{Mayer_2016_CVPR}.
In this work, we follow the former approach, which naturally requires less
parameters, leading to easier to train networks and in our experience also a
better generalization performance.
Especially the latter feature is attractive for our aerial reconstruction task.
Dense ground truth data is notoriously hard to acquire, while artificial datasets
\cite{Mayer_2016_CVPR} usually lack the photogrammetric properties of
'real world scenes' and especially of the specific dataset in consideration.
The problem of missing ground truth data is further magnified by the fact
that CNNs demand a lot of labeled training data to expose their performance. 
While LiDAR measurements could provide at least (very) sparse ground truth, such
an approach would mitigate the advantages of utilizing image based matching at all,
with the additional problem that these measurements appear too sparse to be of
use for CNN training.
Nevertheless, the aim of this work is to utilize CNNs for stereo estimation of
aerial scenes.
To that end, we propose a \emph{self-supervised learning framework}.
Instead of formulating the problem as an unsupervised learning task, which 
ultimately leads to a fully generative approach, we rather
directly utilize the dataset that has
to be reconstructed as training data.
In that sense, we are able to learn the specific imaging characteristics at hand.
Starting from a pre-trained version of our network, we generate the training data
simply by applying our reconstruction method on the whole dataset.
To secure the integrity of our training data we employ strict and conservative
outlier filtering and apply our training procedure on the unmasked,
but still dense data.
%
%
Our experiments indicate that this concept can lead to highly accurate
reconstructions, improving the completeness (and accuracy) 
from 5 (4) percent up to 22 (24) percent, if compared to our pre-trained model
and other competing stereo methods.

%% file: related.tex
Commonly, dense stereo estimation from aerial images is formulated as
a label-based Markov-Random Field (MRF) energy optimization problem, where
methods operate on rectified image pairs.
A popular representative is Semi-Global Matching (SGM)
\cite{Hirschmueller2005,dlr78804} that approximately solves the MRF energy
via dynamic programming (DP), with four scanlines per pixel.
%
Later, the work of Zbontar et al. \cite{Zbontar2016} paved the way for deep learning
for stereo.
They propose to replace the usual, handcrafted features that are
used to define the data term in the energy, with a learned representation.
Later, Luo et al. \cite{Luo2016} exchange the patch-wise training of \cite{Zbontar2016}
with a method that learns the features on whole images instead, introducing a
differentiable cost volume formulation in the CNN.
Both methods rely on SGM to find a solution of their energy formulation and
employ various post-processing steps to refine the solution.
%
Mayer et al. \cite{Mayer_2016_CVPR} instead directly formulate the problem as an
end-to-end regression task. Their CNN possesses several millions of parameters
and, hence, requires a large amount of synthetic data for training.

To overcome the requirement for a sufficient amount of training data, 
the recent trend is to use only weak supervision.
Tonioni et al. \cite{Tonioni_2017_ICCV} generate their training data from a
traditional formulation \cite{Zabih1994}, but estimate a confidence score
for the established matches with another CNN \cite{Poggi_2016_BMVC}.
For training their regression network, the loss function combines the confidence
weight to penalize deviations to their generated training data
with an additional smoothness constraint on the solution.
In contrast, we generate our training data using a state-of-the-art learned
model \cite{knoebelreiter_cvpr2017} and employ a geometrically motivated consistency
check with a hard, conservatively chosen threshold.
%
\cite{Tulyakov_2017_ICCV} explicitly utilize a pre-defined list of matching
constraints to guide the learning. To that end, they are restricted to train
the network per scanline to encode the constraints in the learning procedure.
Another regression based approach is proposed by Zhou et al. \cite{Zhou_2017_ICCV}.
They start from a randomly initialized network and construct their training data
using their own reliable predictions.
Matches are considered as reliable if they survive a left-right (LR)
consistency check.
The network is then trained using only the reliable matches. 
The method is similar in spirit to our approach. In contrast, we advocate to
start from a much better initialization using a pre-trained model
\cite{knoebelreiter_cvpr2017}. In our experience this procedure is both,
beneficial in training time and final accuracy.
Apart from that, our model is much closer to the traditional MRF problem.

%% file: method.tex
In our setting we assume to have access to a larger set of already rectified
image pairs on which we want to perform stereo matching.
What we do not assume is to have access to ground truth data for any of these
image pairs, which could be used for training.
Our objective is to still apply a state-of-the-art stereo CNN and boost its
performance on this specific dataset.
In a nutshell, we exploit a pre-trained and -- during training -- continuously
improving versions of the CNN to generate our own training data.
%

\myparagraph{CNN-CRF Model. }
\label{ssec:cnn-crf}
In this work we utilize the hybrid CNN-CRF model proposed in \cite{knoebelreiter_cvpr2017}
that incorporates deep learning into classical energy minimization.
Our CNN-CRF model minimizes the following typical CRF-type energy defined on
the pixel graph $\mathcal{G}=\{\mathcal{V}, \mathcal{E}\}$ of an image $\Omega$
with the usual 4-connected neighborhood structure $\mathcal{E}$:
\begin{equation}
  \min_{x \in \mathcal{L}} \sum_{i\in \mathcal{V}} f_i(x_i) + \sum_{i\sim j \in \mathcal{E}} f_{ij}(x_i, x_j).
\label{eq:cnn-crf}
\end{equation}
The solution $x^*$ of \eqref{eq:cnn-crf} is a member of the set of mappings
$\mathcal{L}:\mathcal{V}\rightarrow\{0,\ldots,d-1\}^{|\mathcal{V}|}$ representing
a disparity map of $\Omega$ of range $d$.
%
Here, both, the data-term $f_i(x_i)$ and the regularizer $f_{ij}(x_i, x_j)$ are
each represented as a CNN. The optimization of the CRF energy is performed via
a massively parallel and highly efficient variant of dual decomposition. 
The whole system can be learned end-to-end \cite{knoebelreiter_cvpr2017}.
In this work, however, we focus on the data term $f_i$ and keep edge-weights
and penalty function $f_{ij}$ in \eqref{eq:cnn-crf} fix.

\myparagraph{Generating the training data. }
\label{ssec:gt-generation}
To bootstrap our procedure, we directly use the publicly available
model ({\small\url{https://github.com/VLOGroup/cnn-crf-stereo}}) 
with a 7 layered data term CNN,
which was trained on the Middlebury Stereo 2014 dataset \cite{Scharstein2014}.
It has been shown in \cite{knoebelreiter_cvpr2017} that the model
generalizes well to unknown scenes, which arguably makes it a good
candidate for generating our initial training data.
However, because the original training images are completely different from our
aerial dataset, the reconstruction still contains outliers and
erroneous regions.
Therefore, directly using the resulting disparity images for training a new
data term will rather harm the performance than improve our method.
To mitigate this problem, our training procedure has to distinguish between 
regions, where it can trust the generated ground truth and where not.
%
%

\myparagraph{Filtering the generated data.}
\label{ssec:gt-senity-checks}
We use the common left-right consistency check to filter unreliable matches.
Therefore, we first compute two disparity maps, $d_l$ and $d_r$, for each image pair,
where either the left image ($d_l$) or the right one ($d_r$) serves as the
reference frame.
For our filter we then require that matching points in the left
and right image are in mutual correspondence for both disparity maps.
More precisely, a pixel $x$ survives the left-right consistency check if
\begin{equation}
  |d_l(x) + d_r(x + d_l(x))| < \epsilon,
  \label{eq:lr_check}
\end{equation}
where $\epsilon$ is a threshold that is set to $0.9$ in our experiments.
This simple check gets rid of most of the wrong pixels and is,
in our experience, already sufficient to retrain our model.

%
%

\myparagraph{Training. }
\label{ssec:training}
As stated in Section \ref{ssec:cnn-crf}, the model consists of two networks,
one for the data-term $f_i$ and one for the regularizer $f_{ij}$.
From our experience, training edge costs for our regularizer requires the edges
also to be represented in the training data. However, pixel near occlusions
rarely survive our consistency check and are, thus, underrepresented in our
self-supervised training data.
Consequently, we keep the
edge costs fixed and only retrain the network represented by $f_i$ for the aerial images.
In particular, we generate a one-hot encoding of our ground truth disparity maps
and perform maximum likelihood training, i.e. we minimize the
following loss function \wrt the parameters $\theta$ of the network:
\begin{equation}
  L(f(\theta), f^*) \!=\! -\sum_{i \in \Omega} \sum_{d \in D} f_{i,d}^* \log f_{i,d} 
  =
  -\sum_{i \in \Omega} \log f_{i, d^*},
\end{equation}
where $f$ is the correlation volume predicted by the model, $f^*$ is the one-hot
encoding of the ground truth disparity map.
The second equality comes from the fact that the one-hot encoding puts all the
probability mass to the ground truth disparity $d^*$.
%

%% file: experiments.tex
In this section we evaluate the effectiveness of our self learning algorithm
in the context of aerial images.
We compare the depth maps generated by the well-known Semi-Global Matching
algorithm \cite{Hirschmueller2005,libSGM}
with the pre-trained CNN-CRF model and our model, refined via self-supervised training.

\myparagraph{Dataset. }
We evaluate our method on the Vaihingen dataset of the ISPRS Urban classification
and 3D reconstruction benchmark \cite{rottensteiner2013isprs}.
The Vaihingen dataset consists of 20 aerial images of size
$7680 \times 13824$ pixels.
In each image of the dataset the blue channel has been replaced by the response
of an infrared camera, which leads to further deviation between
the pre-trained and refined model.
Nevertheless, we could observe similar behavior for the Toronto dataset of
the same benchmark \cite{rottensteiner2013isprs} where the color channels are RGB.
All images are registered in a global coordinate system. Additionally a
laser point cloud is provided, which we use for our evaluation.
We perform all our experiments at half resolution.
%

Both algorithms, SGM and CNN-CRF, require rectified input images.
%
In order to limit the memory consumption during training, 
we additionally divide the images into parts.
%

\begin{figure*}[tb]
  \centering
  \includegraphics[trim={20px 200px 390px 10px}, clip, scale=0.15, angle=90]{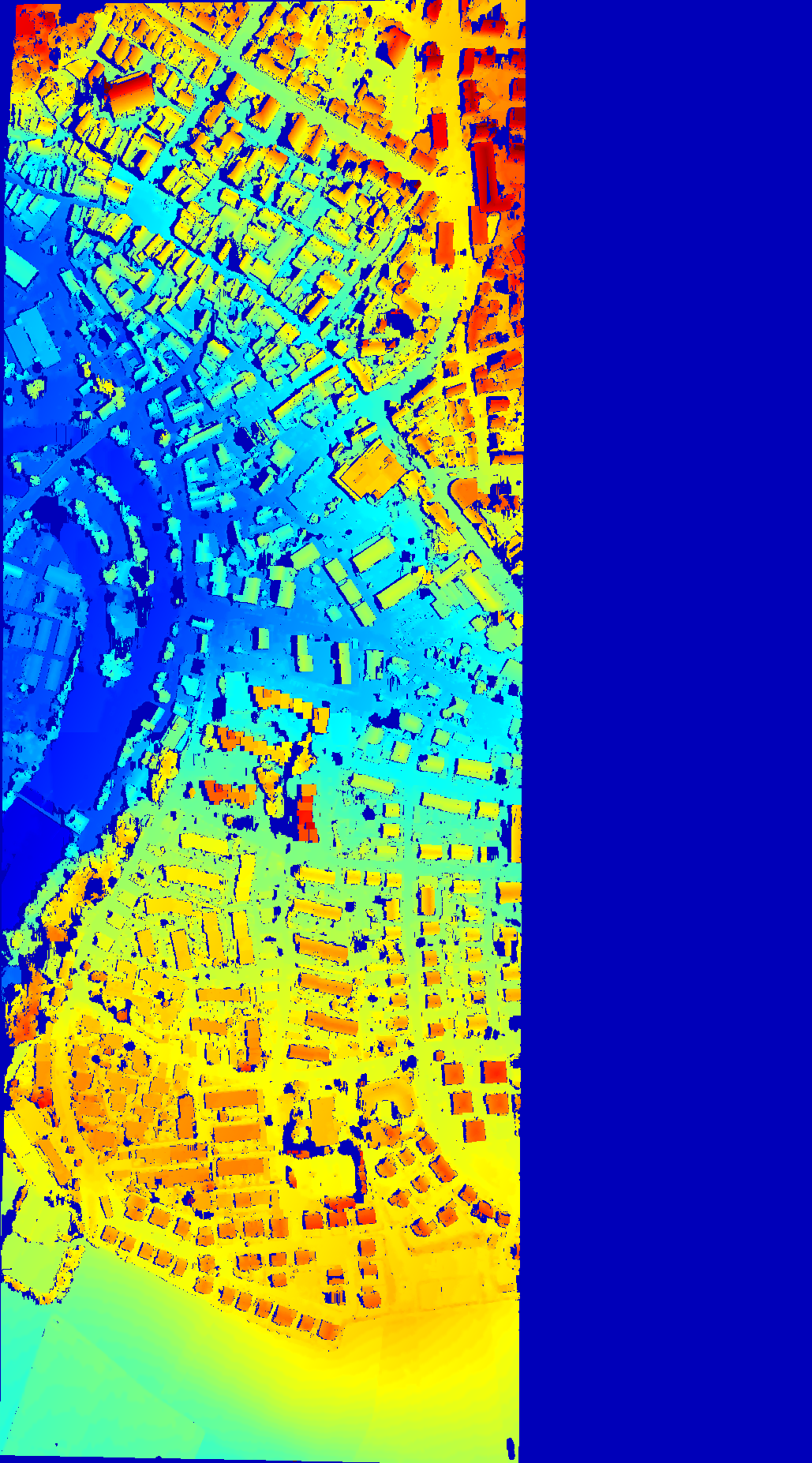} ~
  \includegraphics[trim={20px 200px 390px 10px}, clip, scale=0.15, angle=90]{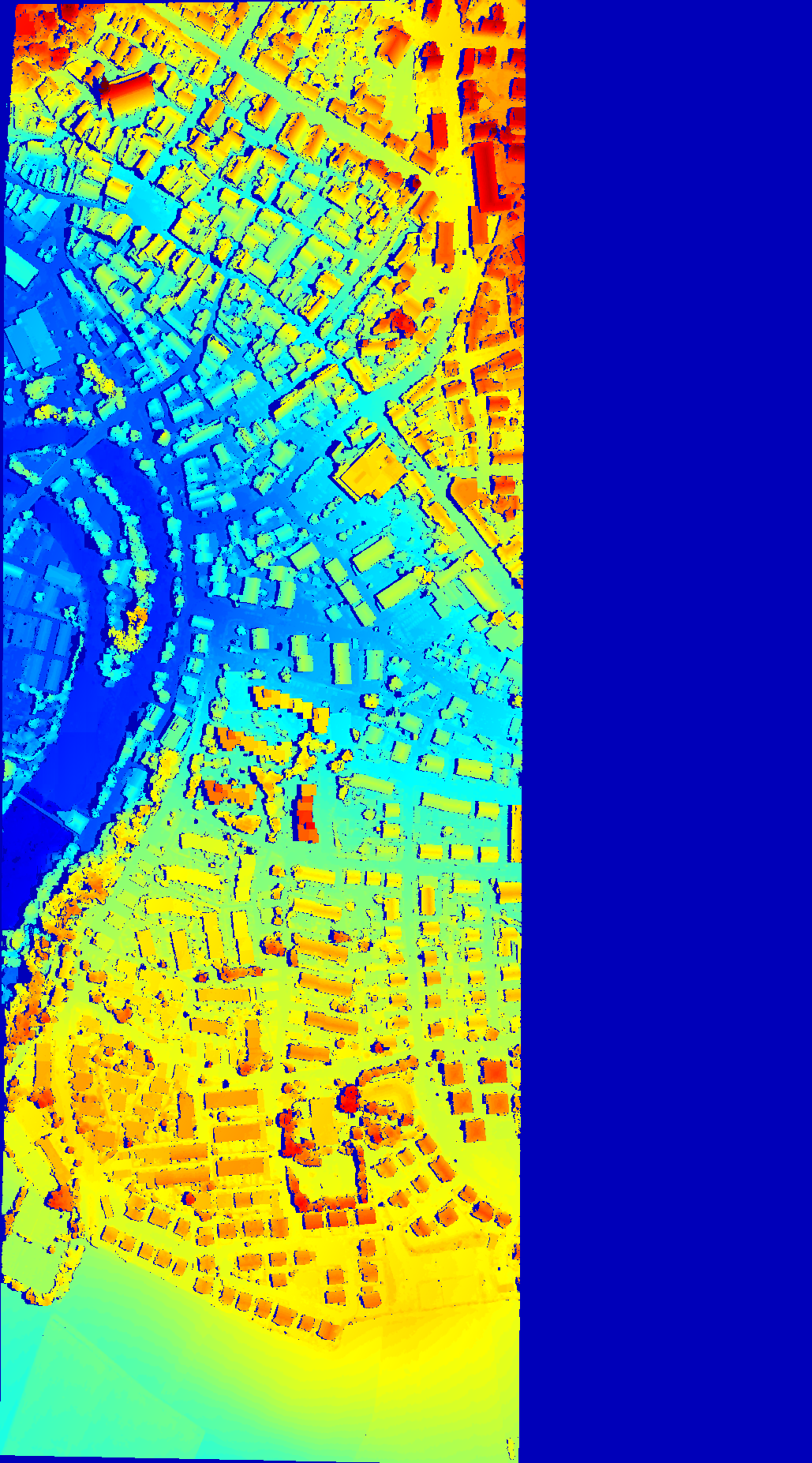}
  \caption{Visual comparison of disparity maps. Left: Generated training data. Right: Improved disparity map after retraining. Most of the (dark blue) artefacts are gone after the self-training. Color-coding from cold (small height) to warm (large height).}
  \label{fig:overview}
  \vspace{-1.2em}
\end{figure*}

\myparagraph{Performance evaluation. }
We use the provided laser scanned depth values as our reference data to compare
the different models.
The pipeline for the evaluation consists of
\RNum{1} computing the disparity map in pixel space for an image pair,
\RNum{2} using the disparity to compute the metric depth value for all pixels in the reference image,
\RNum{3} projecting the laser point cloud into the reference image and
\RNum{4} computing the metric difference for all valid pixels in pixel space.
Additionally, we compute the recall of the
reconstructed points, 
given by
\begin{equation}
  recall = \frac{|\mathcal{P}_S\cap \mathcal{P}_L|}{|\mathcal{P}_L|},
\end{equation}
where $P_S$ is the set of pixel with a valid (surviving the consistency check)
disparity and $P_L$ the set of pixels with a Laser measurement.
A recall of 100\% would mean that every pixel captured by the laser
scanner is also captured by our model.
We perform the evaluation using all available images and, therefore,
report the numbers achieved on the whole dataset.
Recall that we use the laser measurements only for evaluation.

Table \ref{tab:performance} compares the recall and the accuracy
achieved by the baseline SGM model, our pre-trained model used to bootstrap the
training and our model after the first and the second training iteration.
The accuracy is given as the percentage of pixel within a defined 3D distance
to the laser measurements.
In our setting one disparity value corresponds to a 3D displacement
of $0.55$ to $0.72$ meters.
%
Each iteration of the training increases both the recall and the overall accuracy.
Our final model is able to increase the recall by 16.4 percent
points and the accuracy between 2.6 and 12.7 percent points compared to the
pre-trained version.
This shows that self learning is a suitable option to use deep learning
on stereo data without ground truth.
\Fig \ref{fig:overview} visually compares the depth map 
obtained from the pre-trained network with the one
computed with the retrained model. 
Our model is able to close the gaps in the reconstruction 
during retraining.
A closer inspection reveals that masked regions mainly occur near building-ground
edges and correspond to occlusions and, hence, cannot survive the consistency check.
This underlines our findings from Table \ref{tab:performance}, self-training
can improve the accuracy and performance and lead to significantly denser reconstructions.

\begin{table}[tb]
\centering
\begin{tabular}{ccccc}
\toprule
\multirow{2}{*}{\textbf{Model}} & \multirow{2}{*}{\textbf{Recall} [~\%~]} & \multicolumn{3}{c}{\textbf{Accuracy} [~\%~]} \\
& & 0.3m & 0.5m & 1m \\
\midrule
SGM & 76.0 & 52.5 & 69.8 & 86.7 \\
Pt-Net & 87.7 & 62.9 & 76.4 & 87.1\\
Training 1 & 92.1 & \textbf{65.2} & 78.6 & 88.9 \\
Training 2 & \textbf{92.4} & 64.5 & \textbf{78.7} & \textbf{89.3} \\
\bottomrule
\end{tabular}
\caption{Evaluation of the models and comparison with the laser ground truth.
The self-learned models increase the performance on the target domain
significantly compared to SGM and the pre-trained network (Pt-Net) used for retraining.}
\label{tab:performance}
\end{table}

\begin{figure}[tb]
  \centering
  \includegraphics[trim={370px 350px 170px 1000px}, clip, scale=0.17]{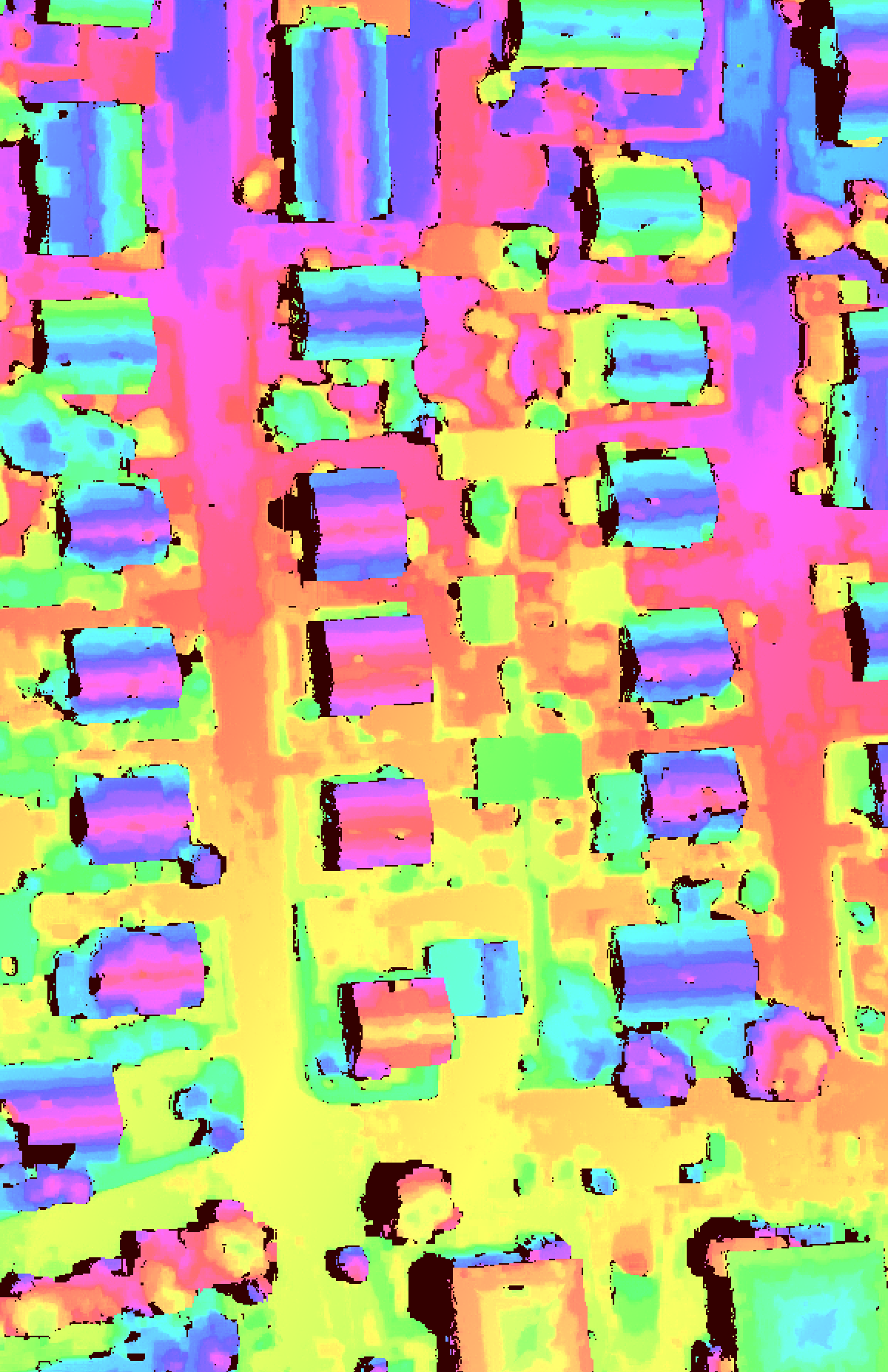}
  \includegraphics[trim={370px 350px 170px 1000px}, clip, scale=0.17]{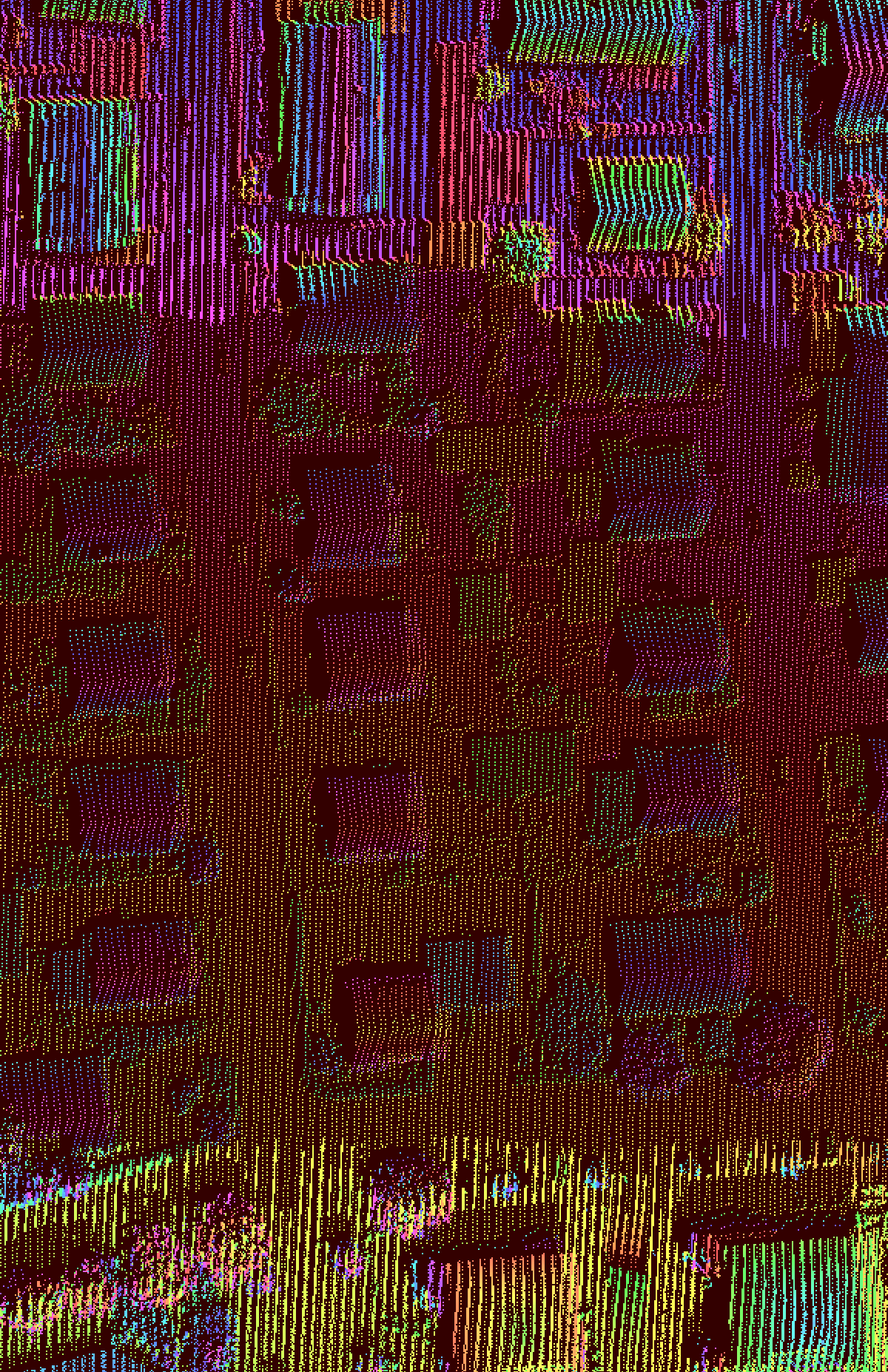}
  \caption{Close-up comparison between our computed depth values and the laser depth values.}
  \label{fig:laser_comparison}
\end{figure}

\begin{figure}[tb]
  \centering
  \includegraphics[trim={0px 0px 0px 0px}, clip, scale=0.11]{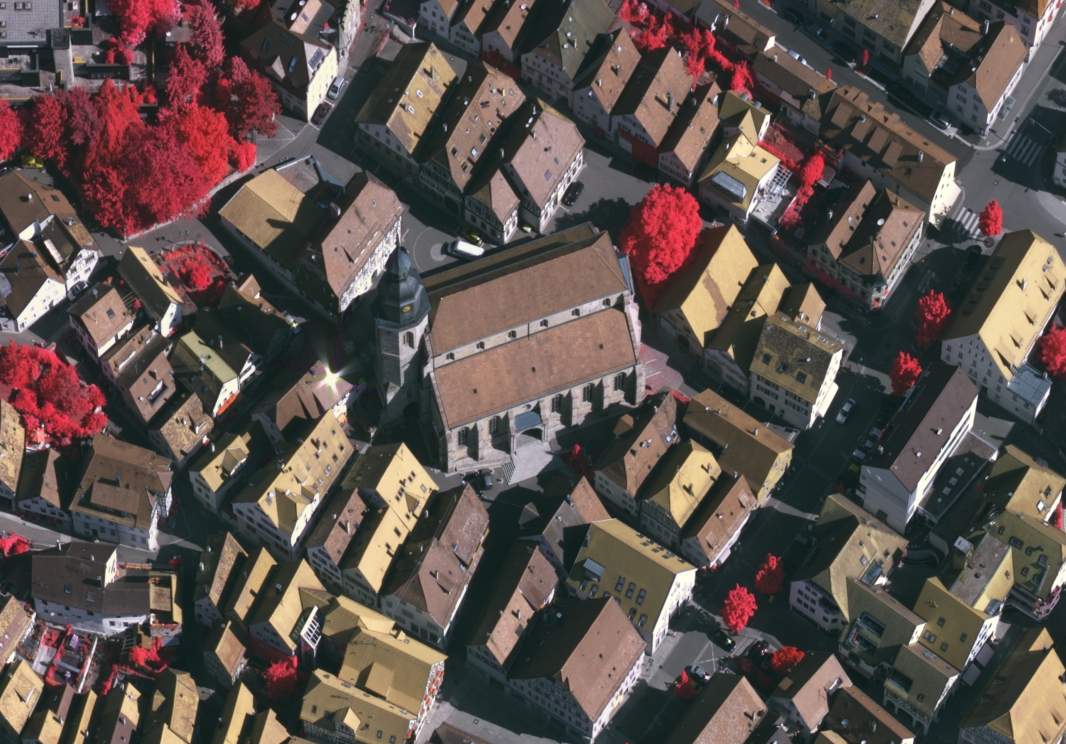}
  \includegraphics[trim={100px 275px 770px 100px}, clip, scale=0.11]{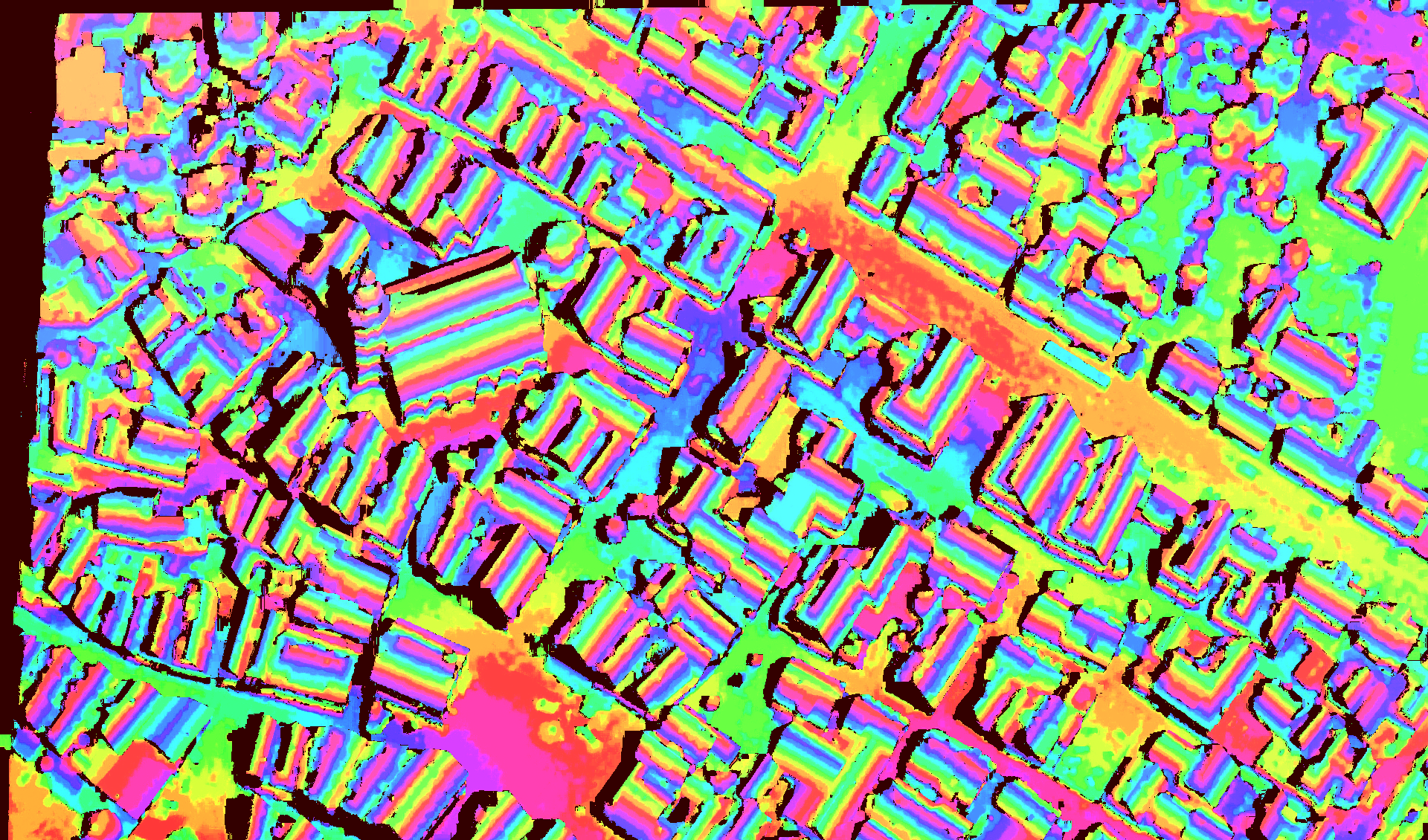}
  \caption{Detailed view of a church. Note how accurately the church tower and the facade is reconstructed.}
  \label{fig:church}
\end{figure}
\Fig \ref{fig:laser_comparison} compares the density of points between our
computed depth map and the laser depth values projected
into the image space.
We are able to densely reconstruct the scene, whereas the laser provides only
a sparse depth map.
\Fig \ref{fig:church} shows a detailed reconstruction of our algorithm.
In this visualization the color-coding is chosen to highlight high-frequency
variations in depth.
Here, especially the tower and the arches in the facade of the church prove
that our model can deliver highly precise reconstruction from aerial images.


%% file: conclusion.tex
We have shown that, without the requirement of any labeled training data,
state-of-the-art machine learning approaches for stereo matching
can be used to compute high quality depth maps from aerial images.
Starting from a pre-trained version, our proposed self-supervised learning
framework constructs the training data with a previous version of the learning
algorithm itself and additionally relies on conservative consistency checking to
reject most of the potential outliers.
%
%
%
Our experiments indicate that this concept works for large scale aerial images,
whose imaging characteristics are quite far from the initial dataset used for
pre-training.
Nevertheless, the perceptual quality as well as the raw performance numbers
are increased significantly compared to baseline models.

